\documentclass{article}
\usepackage{spconf,amsmath,graphicx}

\usepackage{enumitem}
\setlist{nosep, leftmargin=14pt}

\usepackage{mwe} 
\usepackage{multirow}
\usepackage{amsfonts}

\title{Plasma-CycleGAN: Plasma Biomarker-Guided MRI to PET Cross-modality Translation Using Conditional CycleGAN}
\name{
\begin{tabular}{@{}c@{}}
Yanxi Chen$^{1}$, 
Yi Su$^{2}$, 
Celine Dumitrascu$^{3}$, 
Kewei Chen$^{4}$, 
David Weidman$^{2}$,
Richard J Caselli$^{5}$ \\
Nicholas Ashton$^{2}$,
Eric M Reiman$^{2}$,
Yalin Wang$^{1}$
\end{tabular}}

\address{$^{1}$School of Computing and Augmented Intelligence, Arizona State University, Tempe, AZ \\
    $^{2}$Banner Alzheimer’s Institute, Phoenix, AZ, USA
    \quad $^{3}$Basis Scottsdale, Scottsdale, AZ, USA \\
    $^{4}$College of Health Solutions, Arizona State University, Tempe, AZ, USA \\
    $^{5}$Department of Neurology, Mayo Clinic Arizona, Scottsdale, AZ, USA}

\begin{document}

\maketitle
\begin{abstract}
Cross-modality translation between MRI and PET imaging is challenging due to the distinct mechanisms underlying these modalities. Blood-based biomarkers (BBBMs) are revolutionizing Alzheimer's disease (AD) detection by identifying patients and quantifying brain amyloid levels. However, the potential of BBBMs to enhance PET image synthesis remains unexplored. In this paper, we performed a thorough study on the effect of incorporating BBBM into deep generative models. By evaluating three widely used cross-modality translation models, we found that BBBMs integration consistently enhances the generative quality across all models. By visual inspection of the generated results, we observed that PET images generated by CycleGAN exhibit the best visual fidelity. Based on these findings, we propose Plasma-CycleGAN, a novel generative model based on CycleGAN, to synthesize PET images from MRI using BBBMs as conditions. This is the first approach to integrate BBBMs in conditional cross-modality translation between MRI and PET.
\end{abstract}


\section{introduction}
Alzheimer’s disease (AD) is a major neurodegenerative condition affecting millions worldwide, with the number of patients and associated societal costs continually escalating. In the A/T/N classification system for AD diagnosis, brain amyloid, tau pathology, and neurodegeneration are identified via PET scans using various tracers. Among them, amyloid PET scans detect brain amyloid deposition, which signifies an elevated risk of AD clinical symptoms~\cite{jack2016t}.

Despite its accuracy, PET imaging's high cost, radioactivity exposure, and limited availability restrict its widespread use. Alternatively, brain Magnetic Resonance Imaging (MRI) is a non-invasive, widely available tool that detects brain atrophy and has been employed in AD diagnosis. Therefore, synthesizing PET images from MRI scans presents a promising strategy to reduce costs and minimize radiation exposure.

Numerous researchers have utilized advanced image generation algorithms for cross-modality translation, such as GANs and their variants, for this task~\cite{pan2021disease, zhang2022bpgan, zhang2022spatial, hu2021bidirectional, sikka2021mri}. CycleGAN, in particular, has been widely adopted due to its ability to handle unpaired datasets, albeit with some limitations in achieving pixel-wise accuracy~\cite{zhu2017unpaired}. Some noteworthy models for cross-modality translation have been developed based on GAN and CycleGAN. For example, Jin et al. developed BPGAN, which employed gradient profile (GP) loss and structural similarity index measure (SSIM) loss and achieved improvement in SSIM on ADNI dataset~\cite{zhang2022bpgan}. Hu et al. proposed a 3D end-to-end synthesis network named bidirectional mapping GAN (BMGAN) that learns a high-dimension embedding of semantic information of PET images~\cite{hu2021bidirectional}. Recently, the denoising diffusion probabilistic model (DDPM) has emerged as state-of-the-art generative models. Li et al. developed a diffusion model for pathology-aware MRI to PET cross-modality translation (PASTA), which can precisely generate pathology information~\cite{li2024pasta}. Unfortunately, most available methods are based on 2D generative backbones, such as 2D U-Net~\cite{emami2020frea}, 2D cGAN~\cite{shin2020ganbert} and 2D ViT~\cite{zhang2022spatial}. 2D generative models focus on convolution in 2D space, which lacks integrity in one of the dimensions. Therefore, this study focuses on baseline algorithms for standard 3D image inputs using publicly available code. 

Recently, blood-based biomarkers (BBBMs) have emerged as a promising, minimally invasive alternative for early detection of brain amyloid pathology. Available BBBMs mainly include plasma A$\beta$42/40 and phosphorylated Tau family~\cite{janelidze2024plasma}. Notably, the plasma A$\beta$42/40 ratio has demonstrated strong potential in detecting brain amyloid burden and distinguishing AD patients from healthy individuals~\cite{leuzy2022blood, hampel2018blood, chen2024combining}. However, the integration of BBBMs into cross-modality image translation models has not been thoroughly studied.

This study explored the effect of integrating BBBMs into the MRI-to-PET translation models. By conditioning the generative process on the plasma A$\beta$42/40 ratio, we aimed to study whether the image quality of synthesized PET images can be improved. To the best of our knowledge, this is the first study to evaluate the integration of BBBMs into a deep generative framework for cross-modality image translation.

\section{methods}

\subsection{Data Acquisition}
All data used in this study were obtained from the Alzheimer’s Disease Neuroimaging Initiative (ADNI) and were freely available online at the LONI Image and Data Archive (IDA) data repository (https://ida.loni.usc.edu/). We downloaded and processed 1338 image instances of 456 individuals, including 31 AD, 231 mild cognitive impairment (MCI), and 194 cognitively normal (NL) individuals. For each individual, paired MRI and amyloid PET scan results and plasma A$\beta$42/40 ratio from a blood test were available.

\subsection{Data Preprocessing and Augmentation}
Our baseline framework initially processed the input as $256*256*256$ 3-D voxel cubes, encompassing co-registered MRI and PET image pairs. Practically, for more efficient training, we downsampled the images to a size of $128*128*128$ voxels. We implemented a series of data augmentation techniques, including 1) additive Gaussian noise, 2) recursive Gaussian noise, 3) random rotation around each axis, 4) random flip, 5) brightness and contrast variation, and 6) translation. All augmentation methods were under 3D space. During the training cycles, each input image was loaded and went through one or more randomly selected augmentation steps.

\subsection{Baseline Models}

\textbf{Pix2pix Baseline} Pix2pix is a deep generative model for image style translation proposed by Isola et al. in 2017~\cite{isola2017image}. While a GAN model learns a mapping from a random noise vector $z$ to output image $y$: $G:z \rightarrow y$, Pix2pix is based on conditional generative adversarial networks (cGANs), which learns a mapping from the observed image $x$ and a random noise vector $z$ to the output image $y$: $G:\{x, z\} \rightarrow y$  and has been proved to achieve sufficiently good results in multiple tasks. The objective of cGAN can be expressed as: $L_{cGAN}(G, D) = \mathbb{E}_{x, y}[logD(x, y)] + \mathbb{E}_{x, z}[log(1 - D(x, G(x, z)))]$, where $G$ is the generator and $D$ is the discriminator. Pix2pix model also integrates an $L_1$ loss: $L_{L_1}(G) = \mathbb{E}_{x, y, z}[\| y - G(x, y) \|_1]$. The final objective is: $G^* = arg\min_G\max_D L_{cGAN}(G, D) + \lambda L_{L1}(G)$

\textbf{CycleGAN Baseline} CycleGAN was proposed in 2017 and has been widely applied to image translation tasks~\cite{zhu2017unpaired}. a CycleGAN architecture, comprising two generators $G_1$ and $G_2$ and two discriminators $D_1$ and $D_2$. Any network with an encoder-decoder architecture can be used as a generator. CycleGAN facilitates a bidirectional mapping between sMRI and PET domains. Precisely, $G_1$ mapped the MRI domain ($X_M$) to the PET domain ($X_P$) and vice versa for $G_2$, denoted as $G_1: X_M\rightarrow X_P$, and $G_2: X_P\rightarrow X_M$, respectively. Generators $G_1$ and $G_2$ comprised 3-layer 3D CNNs on both encoder and decoder sides, interspersed with 6 Resnet blocks. Each discriminator comprised 5 convolutional layers, followed by 3 fully connected layers. The loss function was: $L(G_1, G_2, D_1, D_2) = L_G(G_1, D_2) + L_G(G_2, D_1) + \lambda_1 L_C(G_1, G_2) + \lambda_2 L_C(G_2, G_1) + \lambda_{idt} L_{idt}$. Here, $L_G$ denoted the generator loss that enforced the similarity between real and fake images through adversarial training against the discriminators. $L_C$ represented the cycle-consistency loss, and $L_{idt}$ was the identity loss that imposed an identity constraint. In our implementation, the learning rate scheduler was replaced by a customized one, and the discriminator was modified following the architecture of ShareGAN~\cite{wang2024joint}.

\textbf{ShareGAN Baseline} ShareGAN is an unsupervised cross-modal synthesis network proposed as a part of a joint learning framework for AD diagnosis~\cite{wang2024joint}. ShareGAN implements an inter-conversion between 3D MRI and PET in a single model. Essentially, ShareGAN has the same architecture as CycleGAN, but the parameters in the two generators are shared. The objective of the ShareGAN model is: $L = L_{GAN} + \lambda_{Cycle}L_{Cycle} + \lambda_{Ide}L_{Ide} + \lambda_{Cls}L_{Cls}$, where both GAN loss $L_{GAN}$ and cycle-consistency loss $L_{Cycle}$ have the same definition as in CycleGAN. Identity loss $L_{Ide}$ forces the synthesis model to achieve an identity mapping if the input is from the output domain, and is defined as: $L_{Ide} = \mathbb{E}_{x_p \in \mathcal{X}_p}\| G_p(x_p) - x_p \|_1 + \mathbb{E}_{x_m \in \mathcal{X}_m}\| G_m(x_m) - x_m \|_1$, where $\mathcal{X}_m$ and $\mathcal{X}_p$ are MRI domain and PET domain, respectively. In this study, we excluded the objective for AD classification by setting $\lambda_{cls} = 0$.

\subsection{Incorporating BBBMs}
\begin{figure}[t]

\begin{minipage}[b]{1.0\linewidth}
  \centering
  \centerline{\includegraphics[width=8.5cm]{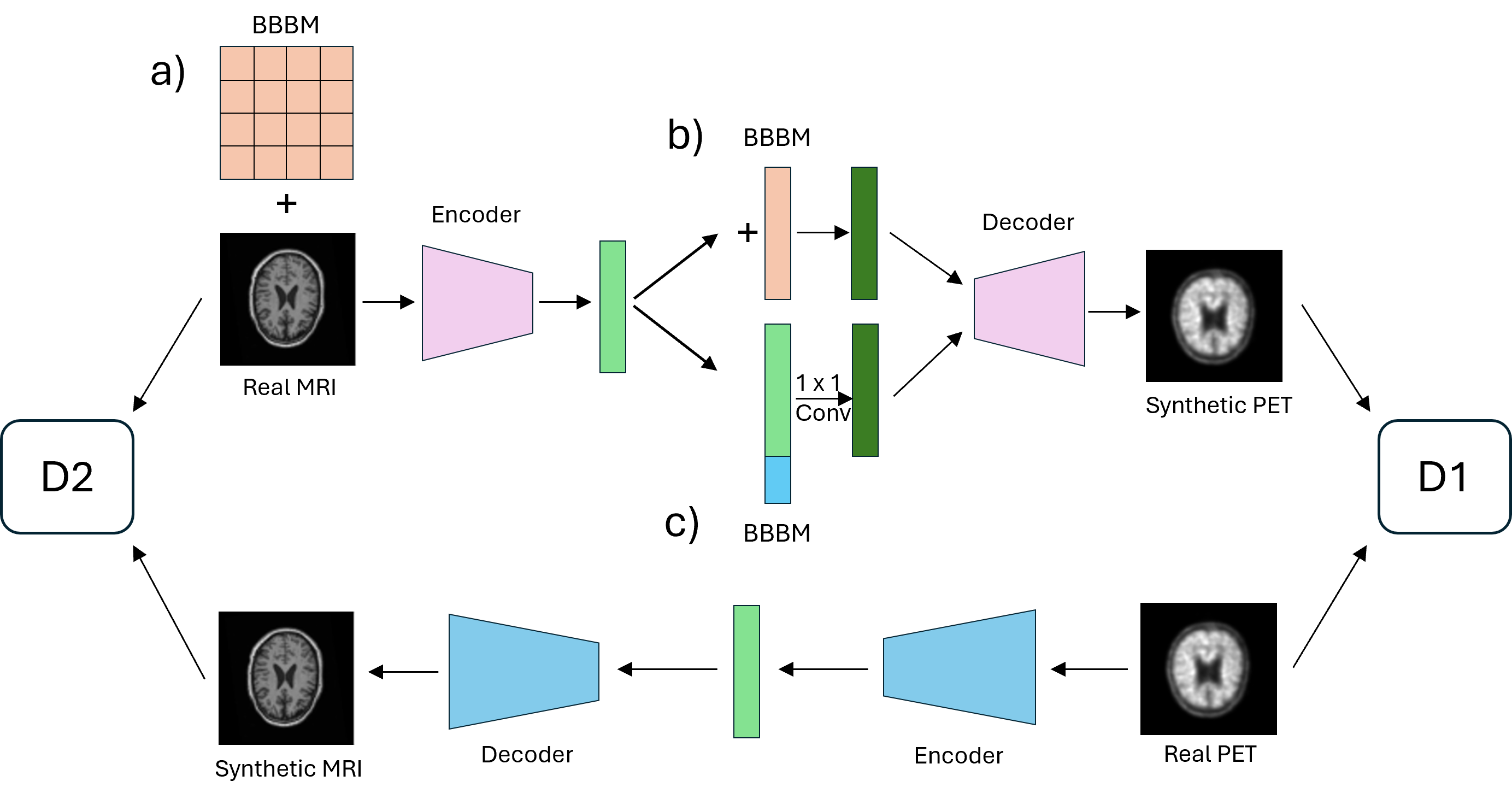}}
\end{minipage}

\caption{Our proposed architecture for BBBM incorporation (using CycleGAN as an example). We implemented three ways to introduce plasma A$\beta$42/40 values into generative frameworks: a) normalized plasma A$\beta$42/40 levels were expanded to the input image size and added to the image. b) normalized plasma A$\beta$42/40 levels were expanded to the size of the latent feature map and added to the feature map. c) normalized plasma A$\beta$42/40 levels were expanded to the size of one channel in latent feature map and concatenated in latent space. $D_1$ and $D_2$ are two discriminators. Pix2pix and ShareGAN have similar architectures, and the BBBM information was added using the same method.}
\label{fig:1}
\end{figure}

We included the BBBM information in our architecture by three methods: 1) expanding normalized plasma A$\beta$42/40 level to the input image size and adding to the image. 2) expanding normalized plasma A$\beta$42/40 level to the size of the latent feature map and adding to the feature map. 3) expanding normalized plasma A$\beta$42/40 level to the size of one channel in latent feature map and concatenating with the feature map in latent space (Fig. 1). Specifically, the size of feature maps after convolution layers was $128*16*16*16$. To integrate BBBM, we took the normalized A$\beta$42/40 levels corresponding to the input images from the clinical data table and expanded the tensor to the same size as the feature map in our bottleneck. The resulting single-channel tensor was concatenated with the feature map along the channel dimension, resulting in a new feature map of size $129*16*16*16$. A 1x1 convolution layer was applied to reduce the feature map’s channel size back to $128*16*16*16$. We applied all three methods for each baseline model and generated the PET images. The performance of all models was evaluated to study whether accounting for additional information alongside structural MRI features improved overall quality and reliability of the generated PET images (Fig. \ref{fig:1}).

\section{experiments and results}

\subsection{Model Training}

With $1338$ paired MRI and FBP PET images, We randomly split the dataset into 910 training, 242 validation and 186 testing. The difference of validation and testing sets was to avoid data leakage. Data leakage was avoided by keeping all images of one individual in the same subset. All models were trained using a learning rate of 0.0002. The coefficients of the loss function were set to $\lambda_1 = \lambda_2 = 10.0$, and $\lambda_{idt} = 0.3$ for CycleGAN and $\lambda_1 = \lambda_2 = 10.0$, and $\lambda_{idt} = 0.5$ for ShareGAN, which achieved best generative quality, respectively. All models were implemented with Pytorch and trained on one Tesla $A100$ GPU for $100$ epochs.

\subsection{Quality Assessment of Generated Images}

\begin{figure*}[t]

\begin{minipage}[b]{1.0\linewidth}
  \centering
  \centerline{\includegraphics[width=1.0\linewidth]{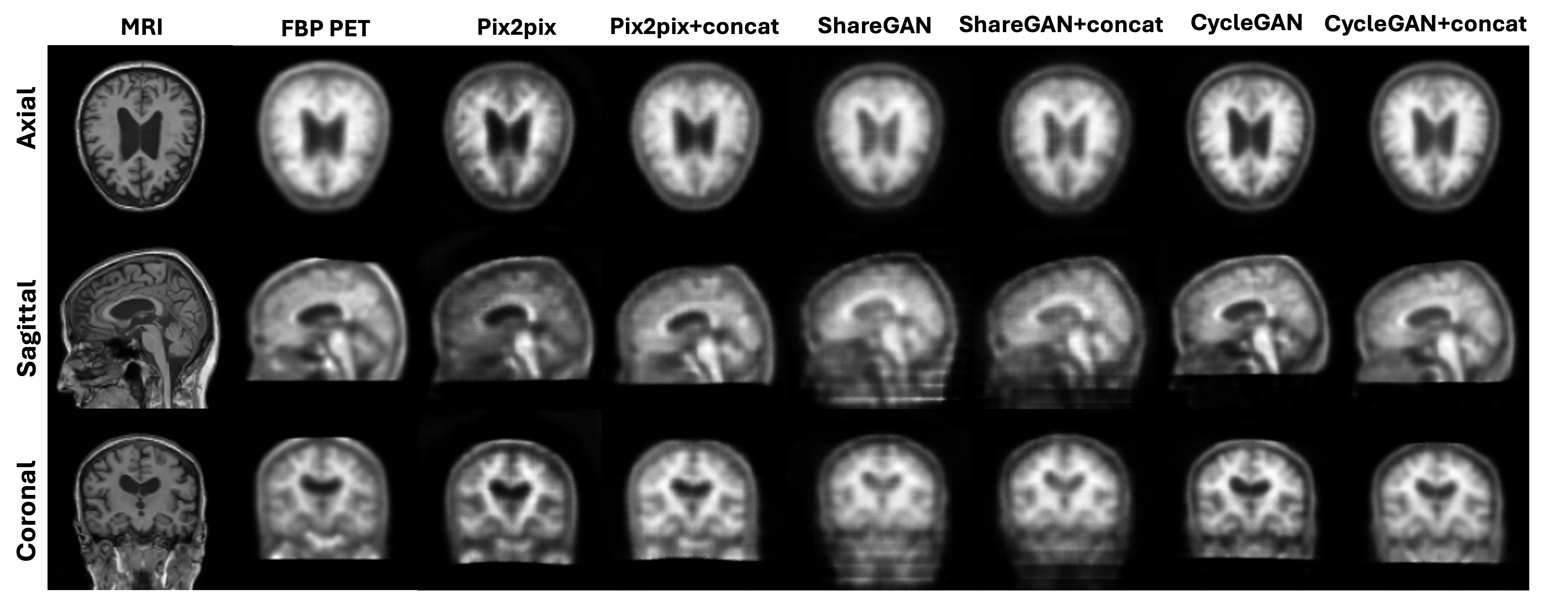}} 
  
\end{minipage}

\caption{Visualization of generate PET images. By visual inspection, CycleGAN generated PET images with more reliable pixel-wise distribution than Pix2pix and ShareGAN. The three rows correspond to axial, sagittal and coronal views, respectively.}
\label{fig:vis}
\end{figure*}

With $186$ amyloid PET images generated from the test set, we evaluated the quality using the most widely used similarity metrics for generative models in computer vision field, including structural similarity index measure (SSIM), peak signal-to-noise ratio (PSNR) and mean squared error (MSE). We evaluated all three baseline models with and without including A$\beta$42/40 as covariate. For each of the baseline models, we tested three ways to incorporate BBBM information, yielding 12 models in total. As a result, including A$\beta$42/40 by concatenating to the latent space consistently improved the performance across all three baseline models, with the highest SSIM being 0.822 for CycleGAN+concat model (Table \ref{table:1}). However, the performance was unstable or decreased in Pix2pix and ShareGAN. In addition, CycleGAN showed the highest robustness against disturbance. Therefore, we only use concatenation in later experiments and named it Plasma-CycleGAN. An example visualization of generated PET images can be seen in Fig. \ref{fig:vis}.

\begin{table}[t]
\centering
\resizebox{\columnwidth}{!}{

\begin{tabular}{c | c | c | c} 
 \hline
 Model & SSIM & PSNR & MSE \\
 \hline
 Pix2pix & 0.766$\pm$0.089 & 23.45$\pm$1.86 & 328.01$\pm$227.53 \\
 Pix2pix+image & 0.683$\pm$0.074 & 22.11$\pm$1.58 & 433.39$\pm$254.33 \\
 Pix2pix+add & 0.465$\pm$0.126 & \textbf{24.60$\pm$1.87} & 
\textbf{259.09.01$\pm$271.76} \\
 \textbf{Pix2pix+concat} & \textbf{0.782$\pm$0.090} & 23.79$\pm$1.97 & 309.29$\pm$262.89 \\
 \hline
 CycleGAN & 0.808$\pm$0.112 & 24.65$\pm$1.90 & 250.33$\pm$174.61 \\
 CycleGAN+image & 0.822$\pm$0.131 & 24.18$\pm$1.64 & 273.70$\pm$217.10 \\
 CycleGAN+add & 0.815$\pm$0.126 & 24.08$\pm$1.96 & 288.26$\pm$237.60 \\
 \textbf{CycleGAN+concat} & \textbf{0.822$\pm$0.113} & \textbf{25.07$\pm$1.82} & 
\textbf{227.00$\pm$182.45} \\
 \hline
 ShareGAN & 0.704$\pm$0.067 & 23.58$\pm$1.48 & 306.18$\pm$162.35 \\
 ShareGAN+image & 0.556$\pm$0.034 & 22.42$\pm$1.33 & 393.77$\pm$177.27 \\
 ShareGAN+add & 0.710$\pm$0.094 & 22.98$\pm$1.23 & 344.14$\pm$155.68 \\
 \textbf{ShareGAN+concat} & \textbf{0.710$\pm$0.065} & \textbf{23.63$\pm$1.41} & \textbf{299.73$\pm$137.62} \\ [0.5pt]

 \hline
\end{tabular}

}

\caption{SSIM, PSNR, and MSE of baseline model and integration method combinations. All models achieved improved performance after incorporating BBBMs in all metrics. Plasma-CycleGAN achieved the highest SSIM, PSNR, and lowest MSE.}

\label{table:1}
\end{table}

\subsection{Standardized Uptake Value Ratio Analysis}
Amyloid positivity is a key marker for Alzheimer’s disease, which refers to a state where the amyloid plaque load in the brain is above a certain threshold. Brain amyloid positivity is usually identified by the masked cerebellum standardized uptake value ratio (MCSUVR). Therefore, in addition to SSIM, PSNR and MSE, measuring similarity of SUVR values between generated and true PET was also crucial. In this study, we conducted two experiments on different SUVR similarity measures, namely SUVR correlation and SUVR classification.

\textbf{SUVR Correlation} To evaluate the model performance with and without BBBM, we calculated the MCSUVR values for each generated image. We assessed the correlation between the generated PET images and the ground truth by calculating the Pearson correlation coefficients (PCCs). As a result, PCCs were greater than 0.5 ($0.770-0.813$), and p-values were significant ($<3e-40$) for all models, which indicated a strong correlation between generated PET and ground truth PET (Table \ref{table:2}). Although the highest PCC was achieved by Pix2pix+concat model ($PCC=0.813$), incorporating BBBM information enhanced the correlation in all models.

\begin{table}[h!]
\centering
\begin{tabular}{c | c | c}
 \hline
 Model & PCC & p-value \\
 \hline
 Pix2pix & 0.786 & 2.83e-40 \\
 \textbf{Pix2pix+concat} & \textbf{0.813} & \textbf{3.57e-45} \\
 \hline
 CycleGAN & 0.777 & 8.53e-39 \\
 \textbf{Plasma-CycleGAN} & \textbf{0.807} & \textbf{5.60e-44}  \\
 \hline
 ShareGAN & 0.770 & 1.01e-37 \\
 \textbf{ShareGAN+concat} & \textbf{0.800} & \textbf{9.13e-43} \\ [0.5pt]
 \hline
\end{tabular}
\caption{PCC and p-values of SUVR correlation between generated and true PET images. All p-values were significant. PCCs of all models were improved by including BBBMs. Although Pix2pix+concat model yielded the highest PCC, all SUVRs correlated strongly with ground truth.}
\label{table:2}
\end{table}

\textbf{SUVR Classification} The diagnosis of preclinical AD is highly dependent on brain amyloid positivity, which can be explicitly calculated from PET images. Therefore, the accuracy of binary classification of amyloid positivity reflects the quality of generated images regarding diagnosis accuracy. We performed binary classification on generated PET images using the same SUVR threshold ($MCSUVR > 1.19$) as in the previous study~\cite{chen2024combining}. As a result, Plasma-CycleGAN achieved the best classification accuracy among all models. However, we observed that the accuracy of Pix2pix and ShareGAN models did not improve after including BBBM (Table \ref{table:3}).

\begin{table}[h!]
\centering
\begin{tabular}{c|c|c} 
 \hline
 Model & Accuracy & F1 \\
 \hline
 Pix2pix & 0.587 & 0.513 \\
 \textbf{Pix2pix+concat} & \textbf{0.772} & \textbf{0.644} \\
 \hline
 CycleGAN & 0.620 & 0.493 \\
 \underline{\textbf{Plasma-CycleGAN}} & \underline{\textbf{0.815}} & \underline{\textbf{0.721}} \\
 \hline
 ShareGAN & 0.609 & 0.514 \\
 \textbf{ShareGAN+concat} & \textbf{0.674} & \textbf{0.571} \\ [0.5pt]
 \hline
\end{tabular}
\caption{Amyloid positivity classification results. Plasma-CycleGAN achieved the best accuracy among all models. No improvement was observed for Pix2pix and ShareGAN.}
\label{table:3}
\end{table}

\section{Discussion and Conclusion}

In this study, we investigated the impact of incorporating BBBM data in generative models for MRI to PET cross-modality translation. This is the first approach to combine BBBMs with generative models for cross-modality translation from MRI to PET. We trained three models with and without including BBBMs on a subset of the ADNI cohort with paired MRI, PET, and plasma A$\beta$42/40 values and evaluated image quality by SSIM, PSNR, and MSE. Our results indicated that including BBBM enhanced the image quality of generated PET images across all models. We computed Pearson correlation coefficients (PCCs) and p-values for the standard uptake value ratios (SUVRs) between the synthesized PET images and the ground truth. As a result, synthesized PET images showed stronger correlations with ground truth SUVR after incorporating BBBM data.

Based on the over quality and stability of the models, we proposed Plasma-CycleGAN, a novel method that is capable of synthesizing PET image from MRI input, conditioned on plasma A$\beta$42/40 level. Future work will investigate the integration of BBBMs into more advanced baselines, such as denoising diffusion probabilistic model (DDPM). In addition, the p-tau family has recently exhibited superior performance over A$\beta$42/40 in AD pathology detection. Specifically, in a recent study, p-tau217 achieved comparable accuracies with cerebrospinal fluid (CSF) biomarkers in brain amyloid and tau detection~\cite{ashton2024diagnostic}. Therefore, with data available in future, we will include other more effective BBBMs in our study.

\section{COMPLIANCE WITH ETHICAL STANDARDS}
This research study was conducted retrospectively using human subject data available in ADNI repository and can be accessed through the LONI Image and Data Archive (IDA) data repository (https://ida.loni.usc.edu/). As confirmed by the license attached to the open-access data, ethical approval was not required.
\section{ACKNOWLEDGMENTS}
This work was supported by grants from the National Institutes of Health (RF1AG073424) and the State of Arizona via the Arizona Alzheimer Consortium. Algorithm development and image analysis for this study were partially supported by the National Institute on Aging (R21AG065942, R01AG069453, RF1AG073424, and P30AG072980), and the State of Arizona via the Arizona Alzheimer Consortium.



\bibliographystyle{IEEEbib}
\bibliography{PlasmaCycleGAN}

\end{document}